\documentclass{article}

\usepackage{arxiv}
\usepackage{xcolor}
\usepackage{caption}
\usepackage{subcaption}
\usepackage{mathtools}
\usepackage{float}
\usepackage{booktabs,siunitx}
\usepackage{makecell}
\usepackage{multirow}
\usepackage[utf8]{inputenc}
\usepackage{booktabs,siunitx}% allow utf-8 input

\usepackage[T1]{fontenc}    % use 8-bit T1 fonts
\usepackage{hyperref}       % hyperlinks
\usepackage{url}            % simple URL typesetting
\usepackage{booktabs}  % professional-quality tables
\usepackage{amsfonts}       % blackboard math symbols
\usepackage{amssymb}
\usepackage{amsthm}
\usepackage{nicefrac}       % compact symbols for 1/2, etc.
\usepackage{microtype}      % microtypography
\usepackage{lipsum}
\usepackage{graphicx}
\usepackage{cite}
\usepackage{amsmath}
\usepackage{color}
\usepackage{cancel}
\usepackage{caption}
\usepackage{subcaption}

\title{EIS - a family of activation functions combining Exponential, ISRU, and Softplus}

\author{
  Koushik Biswas\\%\thanks{Use footnote for providing further information about author (webpage,alternative address)---\emph{not} for acknowledging funding agencies.} 
  Department of Computer Science \\
  IIIT Delhi\\
  New Delhi, India, 110020 \\
  \texttt{koushikb@iiitd.ac.in} \\
   \And
 Sandeep Kumar \\
  Department of Computer Science, IIIT Delhi\\ \& \\
  Department of Mathematics, Shaheed Bhagat Singh College,\\ University of Delhi.\\ 
  New Delhi, India. \\
  \texttt{sandeepk@iiitd.ac.in, sandeep\_kumar@sbs.du.ac.in} \\
   \And
 Shilpak Banerjee \\
  Department of Mathematics \\
  IIIT Delhi\\
  New Delhi, India, 110020 \\
  \texttt{shilpak@iiitd.ac.in} \\
   \And
 Ashish Kumar Pandey \\
  Department of Mathematics\\
  IIIT Delhi\\
  New Delhi, India, 110020 \\
  \texttt{ashish.pandey@iiitd.ac.in} \\
}

\begin{document}
\maketitle

\begin{abstract}
   \iffalse
Neural networks models are found to be one of the most successful models in the era of Artificial intelligence (AI) and deep learning (DL) research. DL models archive state-of-the-art results using activation functions(AF) used in the hidden layers and output layers. AF are essential part of deep neural networks which are widely used to add non-linearity in neural networks. AF can improve training performance and accuracy significantly in Neural Networks. Since past few years of deep learning research, several activation functions has been proposed but Rectified Linear Unit (ReLU) is on top trending list of activation functions. Recently, Swish have been proposed and gain a lot of attention from the research community. We have proposed a novel activation function family, namely EIS with five hyper-parameters, defined as, $\frac{x(\ln(1+e^x))^\alpha}{\sqrt{\beta+\gamma x^2}+\delta e^{-\theta x}}$. We have extracted three simple subfamilies of activations from the above family and they outperformed widely used activation functions. For example, ..........
\fi
Activation functions play a pivotal role in the function learning using neural networks. The non-linearity in the learned function is achieved by repeated use of the activation function. Over the years, numerous activation functions have been proposed to improve accuracy in several tasks. Basic functions like ReLU, Sigmoid, Tanh, or Softplus have been favorite among the deep learning community because of their simplicity. In recent years, several novel activation functions arising from these basic functions have been proposed, which have improved accuracy in some challenging datasets. We propose a five hyper-parameters family of activation functions, namely EIS, defined as, 
\[
\frac{x(\ln(1+e^x))^\alpha}{\sqrt{\beta+\gamma x^2}+\delta e^{-\theta x}}.
\] 
We show examples of activation functions from the EIS family which outperform widely used activation functions on some well known datasets and models. For example, $\frac{x\ln(1+e^x)}{x+1.16e^{-x}}$ beats ReLU by 0.89\% in DenseNet-169, 0.24\% in Inception V3 in CIFAR100 dataset while 1.13\% in Inception V3, 0.13\% in DenseNet-169, 0.94\% in SimpleNet model in CIFAR10 dataset. Also, $\frac{x\ln(1+e^x)}{\sqrt{1+x^2}}$ beats ReLU by 1.68\% in DenseNet-169, 0.30\% in Inception V3 in CIFAR100 dataset while 1.0\% in Inception V3, 0.15\% in DenseNet-169, 1.13\% in SimpleNet model in CIFAR10 dataset.
\end{abstract}

% keywords can be removed
%........................................

\section{Introduction}
Multi-layered neural networks are widely used to learn nonlinear functions from complex data. An activation function is an integral part of neural networks that provides essential non-linearity. A universal activation function may not be suitable for all datasets, and it is important to select an appropriate activation function for the task in hand. Nevertheless, a piecewise activation function, Rectified Linear Unit (ReLU) \cite{relu2,relu1,relu}, defined as $\max(x,0)$, is widely used due to its simplicity, convergence speed and lesser training time. 

Despite of its simplicity and better convergence rate than Sigmoid and Tanh, ReLU has drawbacks like non-zero mean, negative missing, unbounded output, dying ReLU  see\cite{srs} to name a few. Various activation functions have been proposed to overcome the drawbacks of ReLU and improve performance over it. Some of the variants of ReLU are Leaky ReLU \cite{lrelu}, Randomized Leaky Rectified Linear Units (RReLU) \cite{rrelu}, Exponential Linear Unit (ELU) \cite{elu}, Inverse Square Root Linear Units (ISRLUs) \cite{isrelu}, Parametric Rectified Linear Unit (PReLU) \cite{prelu}, and P-TELU \cite{ptelu}. But none of the above-mentioned activation functions have come close to ReLU in terms of popularity. Most recently, Swish \cite{swish} has managed to gain attention from the deep learning community. Swish is a one-parameter family of activation functions defined as ${x \operatorname{sigmoid}(\beta x)}$. Worth noting that, what is popularly recognized by the machine learning community now as the Swish function was first indicated in 2016 as an approximation to the GELU function \cite{gelu}, and again in 2017 was introduced as the SiLU function \cite{silu}, and again for a third time in 2017 as the Swish function \cite{swish}. Though for the time being, we have stuck to the name Swish. Some other hyper-parametrized families of activation functions include Soft-Root-Sign \cite{srs} and TanhSoft \cite{tanhsoft}. In fact, many functions from the TanhSoft family have managed to outperform ReLU and Swish as well.

The most prominent drawback of ReLU is the dying ReLU that is providing zero output for negative input. Many novel activation functions are built to overcome this problem. It was resolved by many activation functions by simply defining a piecewise function that resembles ReLU for positive input and takes non-zero values for negative input. Swish is different from such piecewise activation functions in the sense that it is a product of two smooth functions, and manages to remain close to ReLU for positive input and takes small negative values for the negative input. Recently, a four hyper-parameters family of activation functions, TanhSoft \cite{tanhsoft}, have been proposed and showed that many functions from TanhSoft have the similar closeness to ReLU as Swish, and perform better when compared to ReLU and Swish.

In this work, we have proposed a family of activation functions with five hyper-parameters known as EIS and defined as 
\begin{align}\label{fam}
     \frac{x(\ln(1+e^x))^\alpha}{\sqrt{\beta+\gamma x^2}+\delta e^{-\theta x}}
\end{align}
where $\alpha,\beta,\gamma,\delta$ and $\theta$ are hyper-parameters. 

In the next sections, we have described this family, extracted three subfamilies from (\ref{fam}), and shown that for particular values of hyper-parameters they outperform widely used activation functions, including ReLU and Swish. To validate the performance of the subfamilies, we have given a comprehensive search with different values of hyper-parameters.

%........................................
\section{Related Work}
Several activation functions have been proposed as a substitute to ReLU that can overcome its drawbacks. Because of the dying ReLU problem, it has been observed that a large fraction of neurons become inactive due to zero outcome. Another issue which activation functions face is that during the flow of gradient in the network, the gradient can become zero or diverge to infinity, which are commonly known as vanishing and exploding gradient problems. Leaky Relu \cite{lrelu} has been introduced with a small negative linear component to solve the dying ReLU problem and has shown improvement over ReLU. A hyper-parametric component is incorporated in PReLU \cite{prelu} to find the best value in the negative linear component. Many other improvements have been proposed over the years - Randomized Leaky Rectified Linear Units (RReLU) \cite{rrelu}, Exponential Linear Unit (ELU) \cite{elu}, and Inverse Square Root Linear Units (ISRLUs) \cite{isrelu} to name a few. Swish \cite{swish} is proposed by a team of researchers from Google Brain by an exhaustive search \cite{exs} and reinforcement learning techniques \cite{rll}.

%........................................
\section{EIS Activation Function Family}
In this work, we propose a hyper-parametric family of functions, defined as
\begin{align}
    \mathcal{F}(x;\alpha,\beta,\gamma,\delta,\theta) = \frac{x(\ln(1+e^x))^\alpha}{\sqrt{\beta+\gamma x^2}+\delta e^{-\theta x}}. \label{eq:fam}
\end{align} 
Since family in \eqref{eq:fam} is made up of three basic activation functions - Exponential, ISRU, and Softplus, we name this family as EIS. The family make senses for the hyper-parameter ranges $0 \leq \alpha\leq 1,\ 0 \leq \beta< \infty,\ 0 \leq \gamma< \infty ,\ 0 \leq \delta < \infty$ and $0 \leq \theta < \infty$, and $\beta, \gamma$ and $\delta$ can not simultaneously be equal to zero as this will make the function undefined.
For experimental purposes, we only work with small ranges of hyper-parameters such as $\alpha = 0\ \text{or}\ 1,\ 0 \leq \beta< 3,\ 0 \leq \gamma< 3,\ 0 \leq \delta < 3$ and $0 \leq \theta < 2.5$. The relationship between the hyper-parameters $\beta,\  \gamma,\ \delta,\ \theta$ portrays an important role in the EIS family and controls the slope of the function in both negative and positive axes. The parameter $\alpha$ have been added to switch on and off Softplus function in numerator. For square root, we consider the positive branch. Note that $f(x;0,0,1,0,\theta)$ recover the constant function 1 while $f(x;0,1,0,0,\theta)$ recovers the identity function $x$. Moreover,
\begin{align}
    \lim_{\delta\to\infty}\mathcal{F}(x;0, 0, 1, \delta, 0)= 0 \quad\forall x\in\mathbb{R}.
\end{align}
For specific values of hyper-parameters, EIS family recover some known activation functions. For example,
$\mathcal{F}(x;1,0,1,0,\theta)$ is Softplus \cite{softplus}, $\mathcal{F}(x;0,1,0,1,\theta)$ is Swish \cite{swish}, and $\mathcal{F}(x;0,1,\gamma,0,\theta)$ is ISRU \cite{isrelu} activation functions.

The derivative of the EIS activation family is as follows:
\begin{align}
    &\frac{d}{dx} \mathcal{F}(x;\alpha,\beta,\gamma,\delta,\theta) =\notag\\ &\frac{(\ln(1+e^x))^\alpha}{\sqrt{\beta+\gamma x^2}+\delta e^{-\theta x}} 
    + \frac{\alpha x(\ln(1+e^x))^{\alpha-1}}{\sqrt{\beta+\gamma x^2}+\delta e^{-\theta x}}\ \frac{e^x}{1+e^x} \notag\\ 
    &- \frac{x(\ln(1+e^x))^\alpha}{(\sqrt{\beta+\gamma x^2}+\delta e^{-\theta x})^2}\ \Big(\frac{\gamma x}{\sqrt{\beta+\gamma x^2}} - \delta\theta e^{-\theta x}\Big).
\end{align}

%...........................................

\section{Search Findings}
\begin{figure}[H]
    \begin{minipage}[t]{.49\linewidth}
        \centering
    
        \includegraphics[width=\linewidth]{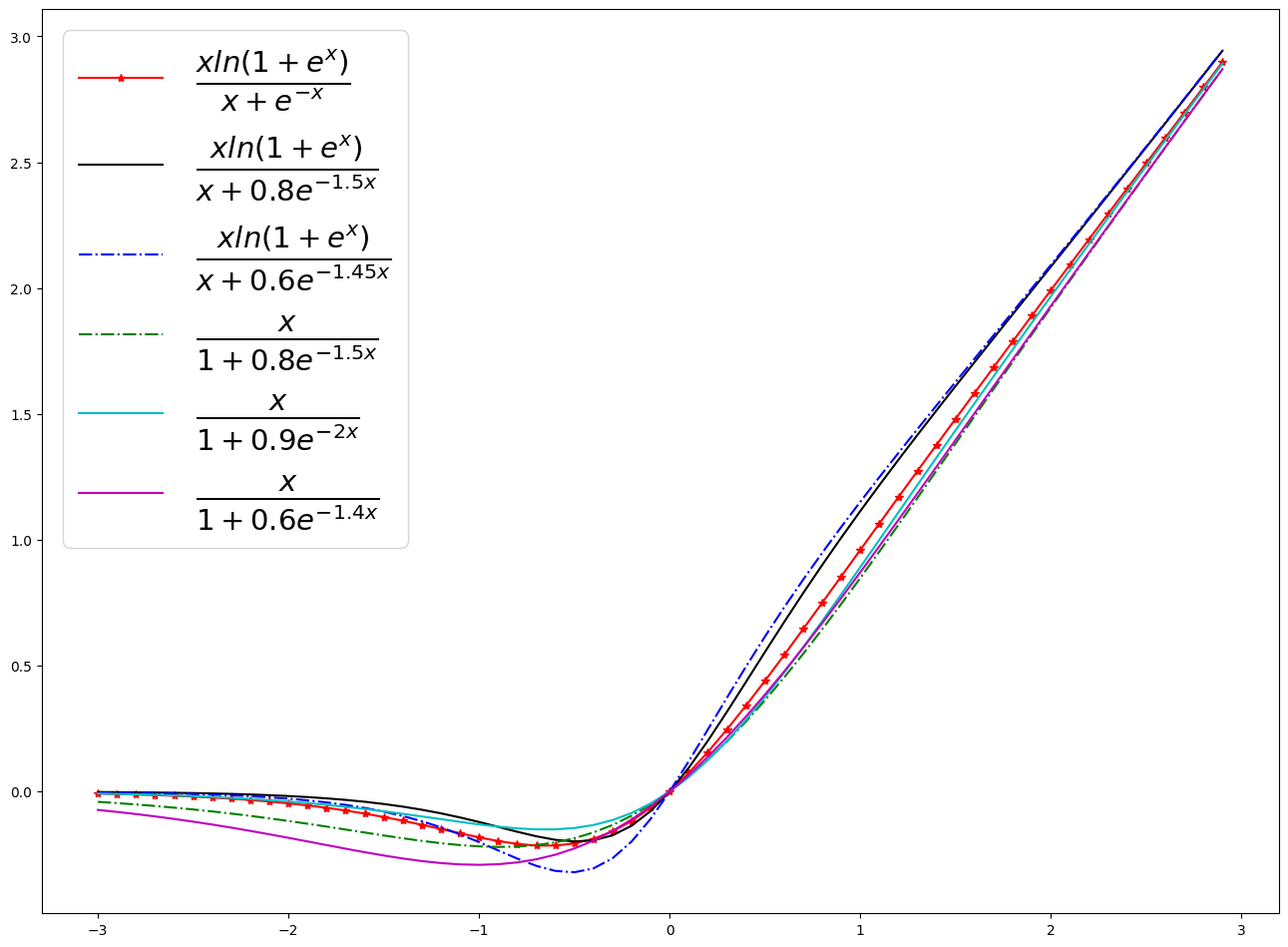}
        
        \label{ser1}
    \end{minipage}
    \hfill
    \begin{minipage}[t]{.49\linewidth}
        \centering
        
       \includegraphics[width=\linewidth]{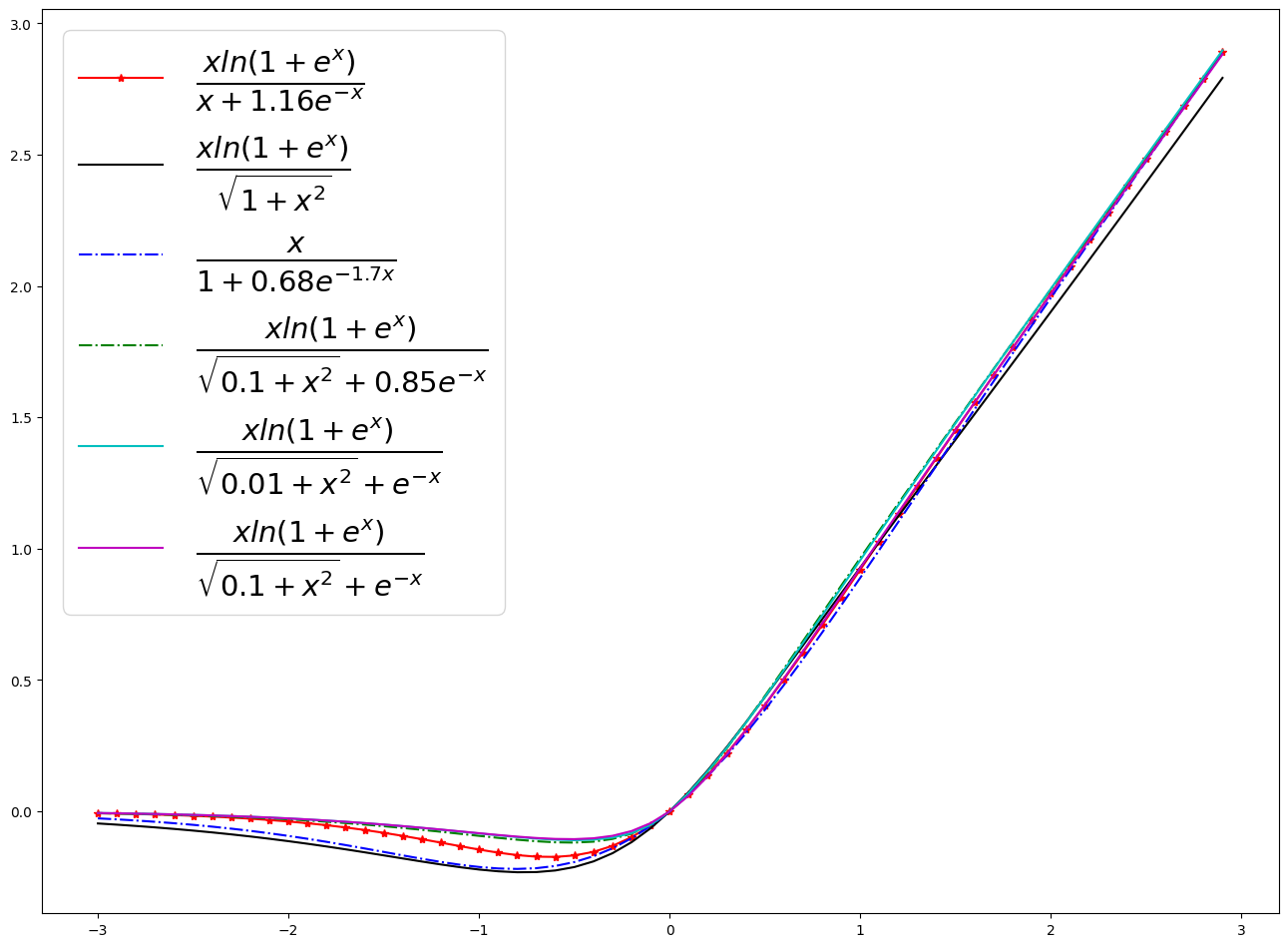}
        
        \label{fig1a}
    \end{minipage}  
    \caption{ A few novel activation functions from the searches of the EIS family.}
\label{ser2}
\end{figure}

We have performed an exhaustive search with EIS family by taking various combinations of hyper-parameters. All of the functions have been trained and tested with CIFAR10 \cite{cifar10} and CIFAR100 \cite{cifar10} datasets with DenseNet-121 (DN-121) \cite{densenet}, MobileNet V2 (MN) \cite{mobile, mobile1}, and SimpleNet (SN) \cite{simple} models and twelve top performing functions have been reported in Table~\ref{tabser} and shown in figure~\ref{ser2}. As evident from the Table~\ref{tabser}, all of these functions have either outperformed or performed equally well when compared with ReLU or Swish. In particular, note that functions $\frac{x\ln(1+e^x)}{x+\delta e^{-\theta x}}$, $\frac{x\ln(1+e^x)}{\sqrt{\beta+\gamma x^2}}$, and $\frac{x}{1+\delta e^{-\theta x}}$ constantly outperform ReLU and Swish. We will discuss these three functions in detail in the next section. Moreover, all functions of the form $\mathcal{F}(x;1,\beta,1,\delta,\theta) = \frac{x\ln(1+e^x)}{\sqrt{\beta+ x^2}+\delta e^{-\theta x}}$ with $0 < \beta\leq 1$, and $0 < \delta \leq 1.5$ and $0<\theta<2$ have shown good performance consistently in our searches but due to higher training time, we have not reported their results with complex models.

%..........................................

%......................................
\begin{table}
\centering
  \begin{tabular}{|l|l|l|l|l|l|l|}
    \hline
    \multirow{2}{*}{Activation Function} &
      \multicolumn{3}{c|}{CIFAR100} &
      \multicolumn{3}{c|}{CIFAR10}\\
      \cline{2-7}
    & \makecell{Top-1\\ accuracy\\ on\\ MN} & \makecell{Top-1\\ accuracy\\on\\ DN-121} &  \makecell{Top-1\\ accuracy\\on\\ SN} & \makecell{Top-1\\ accuracy\\ on\\ MN} & \makecell{Top-1\\ accuracy\\on\\ DN-121} &  \makecell{Top-1\\ accuracy\\on\\ SN} \\
    \hline
    ReLU & 56.90 & 66.20 & 62.77 & 85.49 & 90.69 & 91.07\\[0.15cm]
    \hline
    Swish & 56.25 & 66.91 & 64.85 & 85.55 & 90.80 & 91.70\\[0.15cm]
    \hline
    $\frac{x\ln(1+e^x)}{x+1.16e^{-x}}$
             & 57.24 &  67.45 & 64.99 & 86.63 & 90.99 & 92.01 \\[0.15cm]
            \hline
            $\frac{x\ln(1+e^x)}{\sqrt{1+x^2}}$ & 57.60 & 67.50 & 65.15 & 86.32 & 91.05 &  92.20\\[0.15cm]
            \hline
            $\frac{x}{1+0.68e^{-1.7x}}$ & 57.46 & 67.42 & 65.09 & 86.00 & 91.12  & 92.35\\[0.15cm]
            \hline
            $\frac{x\ln(1+e^x)}{\sqrt{0.1+x^2}+0.85e^{-x}}$ & 57.85 & 67.48 & 65.12 & 86.08 & 91.10 & 92.40\\[0.15cm]
            \hline
            $\frac{x\ln(1+e^x)}{x+e^{-x}}$ & 56.57 & 66.22 & 64.52 & 85.70 & 90.89  & 91.70\\[0.15cm]
            \hline
            $\frac{x\ln(1+e^x)}{x+0.8e^{-1.5x}}$ & 56.03 & 66.03 & 64.12 & 85.36 & 90.50 & 91.93\\[0.15cm]
            \hline
            $\frac{x\ln(1+e^x)}{x+0.6e^{-1.45x}}$ & 56.72 & 66.85 & 64.78 & 85.25 & 90.40  & 91.70\\[0.15cm]
            \hline
            $\frac{x\ln(1+e^x)}{\sqrt{0.1+x^2}+e^{-x}}$ & 57.30 & 67.15 & 65.03 & 85.78 & 90.60 & 91.77\\[0.15cm]
            \hline
            $\frac{x\ln(1+e^x)}{\sqrt{0.01+x^2}+e^{-x}}$ & 57.05 & 67.06 & 64.99 & 86.24 & 90.79 & 92.11\\[0.15cm]
            \hline
            $\frac{x}{1+0.8e^{-1.5x}}$ & 56.69 & 66.25 & 64.77 & 85.36 & 90.69 & 91.93\\[0.15cm]
            \hline
            $\frac{x}{1+0.9e^{-2x}}$ & 57.12 & 67.01 & 65.01 & 85.25 & 90.62 & 91.79\\[0.15cm]
            \hline
            $\frac{x}{1+0.6e^{-1.4x}}$ & 56.75 & 66.77 & 64.82 & 85.61 & 90.42 & 91.65\\[0.15cm]
            \hline
  \end{tabular}
  \vspace{0.5cm}
  \caption{Searches on CIFAR100 and CIFAR10}
        \label{tabser}
\end{table}
%.................................................
It's always difficult to say that an activation function will give the best results on challenging real-world datasets, but from the above searches, it shows that there is a merit that the defined family can generalize and can provide better results compared to widely used activation functions. More detailed results with complex models on more datasets are given in the experiments sections.
%.................................................
\section{EIS-1, EIS-2, and EIS-3}
\begin{figure}[H]
    \begin{minipage}[t]{.31\linewidth}
        \centering
    
        \includegraphics[width=\linewidth]{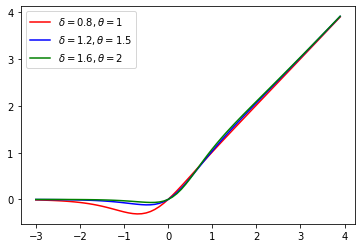}
        \caption{Plots of $\mathcal{F}_1(x;\delta, \theta)$ for different values of $\delta, \theta$.}
        \label{fig11}
    \end{minipage}
    \hfill
    \begin{minipage}[t]{.32\linewidth}
        \centering
    
        \includegraphics[width=\linewidth]{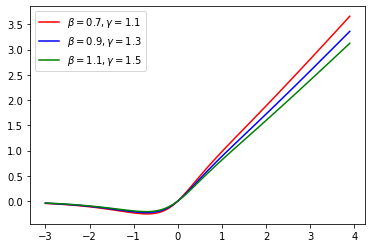}
        \caption{Plots of $\mathcal{F}_2(x;\beta,\gamma)$ for different values of $\beta, \gamma$.}
        \label{fig12}
    \end{minipage}
    \hfill
    \begin{minipage}[t]{.31\linewidth}
        \centering
    
        \includegraphics[width=\linewidth]{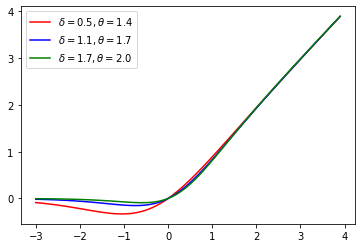}
        \caption{Plots of $\mathcal{F}_3(x;\delta, \theta)$ for different values of $\delta, \theta$.}
        \label{fig1223}
    \end{minipage}
    \end{figure}

\begin{figure}[H]
    \begin{minipage}[t]{.32\linewidth}
        \centering
    
        \includegraphics[width=\linewidth]{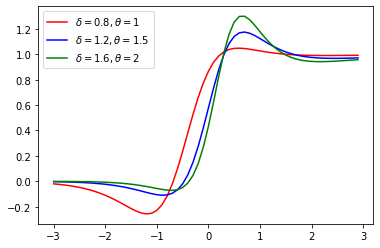}
        \caption{Plots of first derivative of $\mathcal{F}_1(x;\delta, \theta)$ for different values of $\delta, \theta$.}
        \label{figd1}
    \end{minipage}
    \hfill
    \begin{minipage}[t]{.31\linewidth}
        \centering
    
        \includegraphics[width=\linewidth]{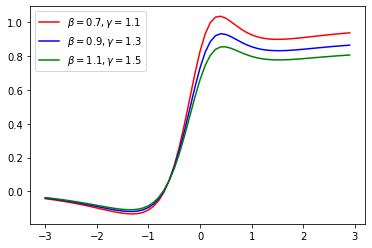}
        \caption{Plots of first derivative of $\mathcal{F}_2(x;\beta,\gamma)$ for different values of $\beta, \gamma$.}
        \label{figd2}
    \end{minipage}
    \hfill
    \begin{minipage}[t]{.31\linewidth}
        \centering
    
        \includegraphics[width=\linewidth]{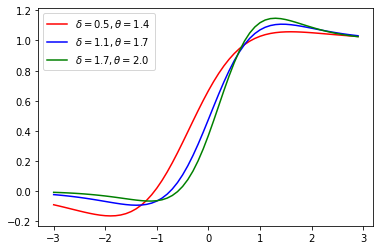}
        \caption{Plots of first derivative of $\mathcal{F}_3(x;\delta, \theta)$ for different values of $\delta, \theta$.}
        \label{figd3}
    \end{minipage}
    \end{figure}

\begin{figure}[H]
    \begin{minipage}[t]{.49\linewidth}
        \centering
    
        \includegraphics[width=\linewidth]{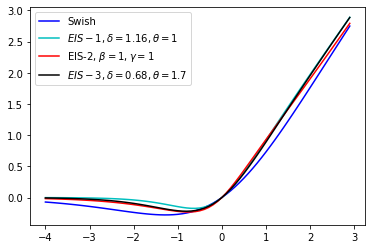}
        \caption{Swish, $\mathcal{F}_1(x;1.16, 1)$, $\mathcal{F}_2(x;1,1)$, and $\mathcal{F}_3(x;0.68,1.7)$}
        \label{com}
    \end{minipage}
    \hfill
    \begin{minipage}[t]{.49\linewidth}
        \centering
        
       \includegraphics[width=\linewidth]{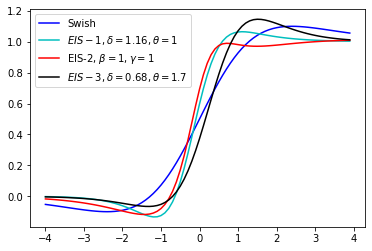}
        \caption{First order derivatives of Swish, $\mathcal{F}_1(x;1.16, 1)$, $\mathcal{F}_2(x;1,1)$, and $\mathcal{F}_3(x;0.68,1.7)$}
        \label{deri}
    \end{minipage}  
\end{figure}

We have extracted three subfamilies of activation functions from the EIS family based on our searches. We call them EIS-1 ($\mathcal{F}_1(x;\delta,\theta)$), EIS-2 ($\mathcal{F}_2(x;\beta,\gamma)$), and EIS-3 ($\mathcal{F}_3(x;\delta,\theta)$). They are defined as follows
\begin{align}
    & \text{EIS-1}:\qquad \mathcal{F}_1(x;\delta,\theta) = \mathcal{F}(x;1,0,1,\delta,\theta) = \frac{x\ln(1+e^x)}{x +\delta e^{-\theta x}},\\
    & \text{EIS-2}:\qquad \mathcal{F}_2(x;\beta,\gamma) = \mathcal{F}(x;1,\beta,\gamma,0,\theta) = \frac{x\ln(1+e^x)}{\sqrt{\beta+\gamma x^2}},\\
    & \text{EIS-3}:\qquad \mathcal{F}_3(x;\delta,\theta) = \mathcal{F}(x;0,1,0,\delta,\theta) =\frac{x}{1+\delta e^{-\theta x}}.
\end{align}
The derivative of the above subfamilies are

\begin{align}
    & \text{Derivative of EIS-1}:\qquad \frac{d}{dx}\mathcal{F}_1(x;\delta,\theta) = \frac{\ln(1+e^x)}{x +\delta e^{-\theta x}}+\frac{x}{x +\delta e^{-\theta x}}\ \frac{e^x}{1+e^x}-\frac{(1-\delta\theta e^{-\theta x})(x\ln(1+e^x))}{(x +\delta e^{-\theta x})^2},\\
    & \text{Derivative of EIS-2}:\qquad \frac{d}{dx}\mathcal{F}_2(x;\beta,\gamma) = \frac{\ln(1+e^x)}{\sqrt{\beta+\gamma x^2}}+\frac{x}{\sqrt{\beta+\gamma x^2}}\frac{e^x}{1+e^x}-\frac{\gamma x^2 \ln(1+e^x)}{(\beta+\gamma x^2)^{\frac{3}{2}}},\\
    & \text{Derivative of EIS-3}:\qquad \frac{d}{dx}\mathcal{F}_3(x;\delta,\theta) =\frac{1}{1+\delta e^{-\theta x}}+\frac{\delta \theta xe^{-\theta x}}{(1+\delta e^{-\theta x})^2}.
\end{align}
%......................................
Graph of some functions from these three families are given in Figures~\ref{fig11}, \ref{fig12}, and \ref{fig1223}. The first order derivatives of these functions are shown in Figures~\ref{figd1}, \ref{figd2}, and \ref{figd3}. Moreover, one function from each of these three families and their derivatives are compared with Swish in Figures~\ref{com} and \ref{deri}. As evident from plots, chosen functions of these three subfamilies have bounded negative domain, smooth derivative and, non-monotonic curve like Swish.

%------------------------------------

%.................................................
%.................................................

%...................................................

\section{Experiments with EIS-1, EIS-2, and EIS-3}
We repeated our search on these three subfamilies and found that  $\mathcal{F}_1(x;1.16,1)$, $\mathcal{F}_2(x;1,1)$, and $\mathcal{F}_3(x;0.68,1.7)$ constantly outperform other functions. In what follows, we will discuss the results of these three functions in more detail. We have included the following baseline activation functions for our comparison.

\begin{itemize}
    \item \textbf{ Rectified Linear Unit (ReLU):-} ReLU was proposed by Nair and Hinton (\cite{relu2},\cite{relu1},\cite{relu}) and currently it is one of the most frequently used activation function in deep learning field. ReLU is defined as 
\begin{align}
    \operatorname{ReLU}(x) = \max(0,x).
\end{align}
%.............................
\item \textbf{Leaky Rectified Linear Unit:-}
To overcome the drawbacks of ReLU, Leaky ReLU was introduced by Mass et al. in 2013 \cite{lrelu} and it shows promising results in different real world datasets. Leaky ReLU is defined as 
\begin{align} 
\operatorname{Leaky ReLU}(x)& = 
    \begin{cases}
       x & x > 0 \\               
      0.01x & x\leq 0
    \end{cases}.
\end{align}

%................
\item \textbf{Exponential Linear Units:-}
ELU is defined in such a way so that it overcomes the vanishing gradient problem of ReLU. ELU was introduced by Clevert et al.
in 2015 \cite{elu}. ELU is defined as 
\begin{align} 
\operatorname{ELU}(x)& = 
    \begin{cases}
       x & x > 0 \\               
      \alpha (e^x-1) & x\leq 0
    \end{cases}
\end{align}
where $\alpha$ is a hyper-parameter. 
%...............................
\item \textbf{Softplus:-}
Softplus \cite{softplus, softplus1} is a smooth non-monotonic function, defined as 
\begin{align}
    \operatorname{Softplus}(x) = \ln(1+e^x).
\end{align}
%................................

\item \textbf{Swish:-}
Swish is a non-monotonic, smooth function which is bounded below and unbounded above \cite{swish}. Swish is defined as 
\begin{align}
    \operatorname{Swish}(x) = x \sigma(x) = \frac{x}{1+e^{-x}}.
\end{align}
\end{itemize}

%.......................................

\begin{figure}[H]
    \centering
    \includegraphics[width=7cm,height=5cm]{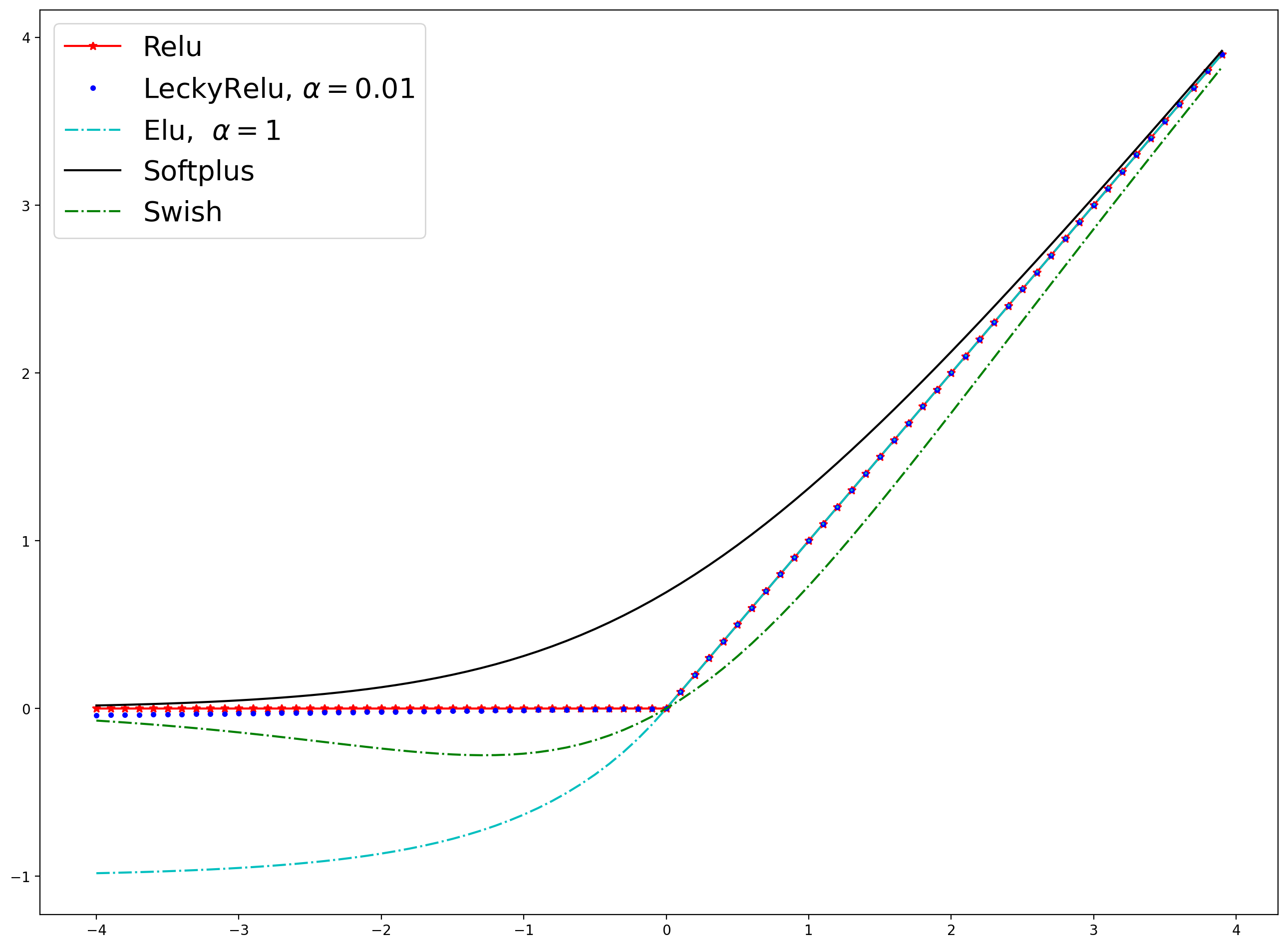}
    \caption{Plots of few widely used activation functions.}
    \label{fig:all122}
\end{figure}

 %.........................................

Table~\ref{tab49} provides a detailed comparison of $\mathcal{F}_1(x;1.16,1)$, $\mathcal{F}_2(x;1,1)$ and $\mathcal{F}_3(x;0.68,1.7)$ with baseline activation functions we compared in our experiments. For comparison, we have used different models such as DenseNet(DN) \cite{densenet}, MobileNet V2(MN) \cite{mobile}, Inception Module V3(IN) \cite{incep}, SimpleNet(SN) \cite{simple} and WideResNet(WRN) \cite{wrn}. We have reported a summary in Table~\ref{tab49} which shows the total number of models for which these three functions beats, equal or underperforms the baseline functions considered for our experiments. 

%.........................................
%-----------------------------
\begin{table}[H]
\newenvironment{amazingtabular}{\begin{tabular}{*{50}{l}}}{\end{tabular}}
\centering
\begin{amazingtabular}
\midrule
Baselines & ReLU & Leaky ReLU & ELU & Swish & Softplus\\
\midrule
$\mathcal{F}_1(x;1.16,1)$ > Baseline & \hspace{0.3cm}14 & \hspace{0.45cm}13 & \hspace{0.3cm}14 & \hspace{0.3cm}14 & \hspace{0.3cm}14\\
$\mathcal{F}_1(x;1.16,1)$ = Baseline & \hspace{0.3cm}0 & \hspace{0.45cm}0 & \hspace{0.3cm}0 & \hspace{0.3cm}0 & \hspace{0.3cm}0\\
$\mathcal{F}_1(x;1.16,1)$ < Baseline & \hspace{0.3cm}0 & \hspace{0.45cm}1 & \hspace{0.3cm}0 & \hspace{0.3cm}0 & \hspace{0.3cm}0\\
\midrule
$\mathcal{F}_2(x;1,1)$ > Baseline & \hspace{0.3cm}14 & \hspace{0.45cm}14 & \hspace{0.3cm}14 & \hspace{0.3cm}14 & \hspace{0.3cm}14\\
$\mathcal{F}_2(x;1,1)$ = Baseline & \hspace{0.3cm}0 & \hspace{0.45cm}0 & \hspace{0.3cm}0 & \hspace{0.3cm}0 & \hspace{0.3cm}0\\
$\mathcal{F}_2(x;1,1)$ < Baseline & \hspace{0.3cm}0 & \hspace{0.45cm}0 & \hspace{0.3cm}0 & \hspace{0.3cm}0 & \hspace{0.3cm}0\\
\midrule
$\mathcal{F}_3(x;0.68,1.7)$ > Baseline & \hspace{0.3cm}14 & \hspace{0.45cm}13 & \hspace{0.3cm}14 & \hspace{0.3cm}14 & \hspace{0.3cm}14\\
$\mathcal{F}_3(x;0.68,1.7)$ = Baseline & \hspace{0.3cm}0 & \hspace{0.45cm}0 & \hspace{0.3cm}0 & \hspace{0.3cm}0 & \hspace{0.3cm}0\\
$\mathcal{F}_3(x;0.68,1.7)$ < Baseline & \hspace{0.3cm}0 & \hspace{0.45cm}1 & \hspace{0.3cm}0 & \hspace{0.3cm}0 & \hspace{0.3cm}0\\
\midrule
\end{amazingtabular}
\vspace{0.5cm}
  \caption{Baseline table of $\mathcal{F}_1(x;1.16,1)$, $\mathcal{F}_2(x;1,1)$ and $\mathcal{F}_3(x;0.68,1.7)$ for Top-1 Accurecy}
  \label{tab49}
\end{table}

%.................................................
\section*{Implementation}
Activation functions $\mathcal{F}_1(x;1.16,1)$, $\mathcal{F}_2(x;1,1)$ and $\mathcal{F}_3(x;0.68,1.7)$ have been tested on different CNN architectures. Models and activations functions are implemented using Pytorch \cite{pytorch}, TensorFlow \cite{tensorflow} and Keras \cite{keras} platforms. We have implemented a seven-layer custom model to evaluate the performance on MNIST, Fashion MNIST, and SVHN datasets and trained with a uniform 0.001 learning rate, Glorot normal initializer \cite{gnormal}, batch normalization \cite{batch} and Adam \cite{adam} optimizer. Also, on CIFAR10 and CIFAR100 datasets, DenseNet, MobileNet V2, and Inception Module V3, WideResNet are trained with uniform 0.001 learning rate, Adam optimizer, batch normalization till 100 epochs. With similar settings, SimpleNet model trained with 200 epochs to evaluate performance on CIFAR10 and CIFAR100 datasets.
%...............................................
\subsection*{Database}
We have reported results with five bench-marking databases, MNIST, Fashion MNIST, The Street View House Numbers, CIFAR10, and CIFAR100. A brief description about the databases are as follows. 
\begin{itemize}
\item \textbf{MNIST}:- MNIST \cite{mnist} is a  well established standard database consisting of 28 $\times$ 28 pixels grey-scale images of handwritten digits from 0 to 9. It is widely used to establish the efficacy of various deep learning models. The dataset consists of 60k training images and 10k testing images.
\item \textbf{Fashion-MNIST}:- Fashion-MNIST \cite{xiao2017fashion} is a database consisting of 28 $\times$ 28 pixels grey-scale images of Zalando's, ten fashion items class like  T-shirt, Trouser, Coat, Bag, etc.  It's consist of 60k training examples and 10k testing examples. Fashion-MNIST provides a more challenging classification problem than MNIST and seeks to replace MNIST. 
\item \textbf{The Street View House Numbers (SVHN) Database}:- SVHN \cite{SVHN} is a popular computer vision database consists of real world house number with $32\times 32$ RGB images. The database has 73257 training images and 26032 testing images. The database has a total of 10 classes.
\item \textbf{CIFAR}:- The CIFAR \cite{cifar10} (Canadian Institute for Advanced Research), is another standard well established computer-vision dataset that is generally used to establish the efficacy of deep learning models. It contains 60k color images of size 32 $\times$ 32, out of which 50k are training images, and 10k are testing images. It has two versions CIFAR 10 and CIFAR100, which contains 10 and 100 target classes, respectively.  
\end{itemize}
%...................................................
\section*{Evaluation of EIS}
\begin{figure}[H]
    \begin{minipage}[t]{.485\linewidth}
        \centering
    
        \includegraphics[width=\linewidth]{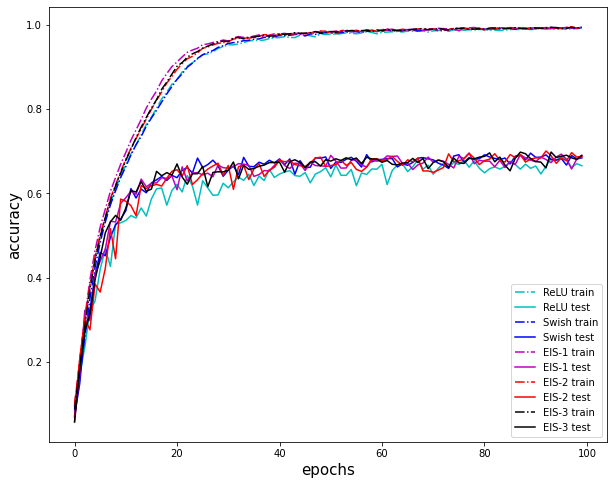}
        \caption{Train and Test accuracy on CIFAR100 dataset with WideResNet 28-10 model}
        \label{acc}
    \end{minipage}
    \hfill
    \begin{minipage}[t]{.48\linewidth}
        \centering
        
       \includegraphics[width=\linewidth]{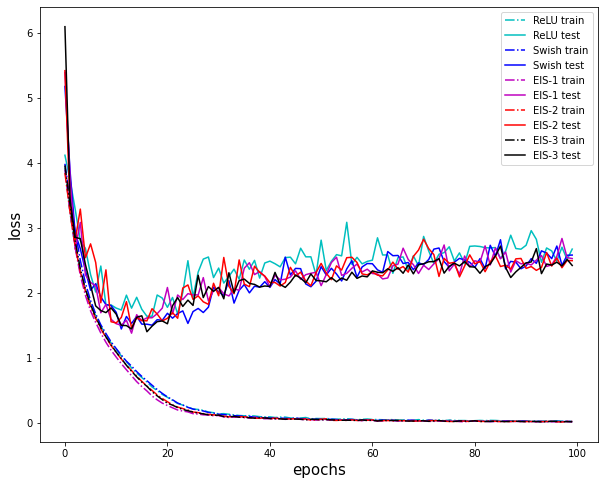}
        \caption{Train and Test loss on CIFAR100 dataset with WideResNet 28-10 model}
        \label{loss}
    \end{minipage}  
\end{figure}

All three activation functions are compared with state-of-the-art five baseline activation functions. The test accuracy of five bench-marking databases have been reported in table~\ref{tab1}, \ref{tab122}, \ref{tab123}, \ref{tab22} and \ref{tab23}. A accuracy and loss plot on WRN 28-10 model with CIFAR100 dataset for $\mathcal{F}_1(x;1.16,1)$, $\mathcal{F}_2(x;1,1)$, $\mathcal{F}_3(x;0.68,1.7)$, ReLU and Swish is given in Figures~\ref{acc} and \ref{loss}. 
%...................................................

\begin{table}[H]
\begin{minipage}{.5\linewidth}
      
\centering
\begin{tabular}{ |c|c|c| }
 \hline
 Activation Function &  \makecell{5-fold mean\\ Accuracy on\\ MNIST data}  \\
 \hline
 $\mathcal{F}_1(x;1.16,1)$ &  99.30\\ 
 \hline
 $\mathcal{F}_2(x;1,1)$ &  99.32\\ 
 \hline
 $\mathcal{F}_3(x;0.68,1.7)$ &  \textbf{99.40}\\ 
 \hline
 ReLU  &  99.17 \\ 
 \hline
 Swish  & 99.21 \\
 \hline
 Leaky ReLU($\alpha$ = 0.01) & 99.18\\
 \hline
 ELU ($\alpha = 1$) &  99.15\\
 \hline
 Softplus & 99.02 \\
 \hline
 \end{tabular}
 \vspace{0.2cm}
\caption{Experimental Results with MNIST Dataset} 
\label{tab1}
\end{minipage}
\begin{minipage}{.5\linewidth}
\centering
\begin{tabular}{ |c|c|c| }
 \hline
 Activation Function &  \makecell{5-fold mean\\ Accuracy on\\ Fashion MNIST data}  \\
 \hline
 $\mathcal{F}_1(x;1.16,1)$ &  \textbf{93.35}\\ 
 \hline
 $\mathcal{F}_2(x;1,1)$ & 93.15 \\ 
 \hline
 $\mathcal{F}_3(x;0.68,1.7)$ &  93.20\\ 
 \hline
 ReLU  &  92.85 \\ 
 \hline
 Swish  & 92.97 \\
 \hline
 Leaky ReLU($\alpha$ = 0.01) & 92.91\\
 \hline
 ELU ($\alpha = 1$) &  92.80\\
 \hline
 Softplus &  92.30\\
 \hline
 \end{tabular}
 \vspace{0.2cm}
\caption{Experimental Results with Fashion MNIST Dataset} 
\label{tab122}

\end{minipage}
\end{table}
%............................................
\begin{table}[H]
\begin{center}
\begin{tabular}{ |c|c|c| }
 \hline
 Activation Function &  \makecell{5-fold mean Accuracy on SVHN data  }\\
 \hline
 $\mathcal{F}_1(x;1.16,1)$ &  95.38\\ 
 \hline
 $\mathcal{F}_2(x;1,1)$ &  95.30\\ 
 \hline
 $\mathcal{F}_3(x;0.68,1.7)$ &  \textbf{95.41}\\ 
 \hline
 ReLU  &   95.20\\ 
 \hline
 Swish  & 95.23\\
 \hline
 Leaky ReLU($\alpha$ = 0.01) & 95.22 \\
 \hline
 ELU($\alpha = 1$)  &  95.20\\
 \hline
 Softplus & 95.10 \\
 \hline
 \end{tabular}
 \vspace{0.2cm}
\caption{Experimental Results with SVHN Dataset} 
\label{tab123}
\end{center}
\end{table}
%...........................................

\begin{table}[H]
\begin{center}
\begin{tabular}{ |c|c|c|c|c|c|c|c|c|c|c| }
 \hline
\makecell{Activation\\ Function} &  \makecell{DN-121\\
Top-1 \\ Acc.}
& \makecell{DN-121\\ Top-3 \\ Acc.}
& \makecell{DN-169\\
Top-1\\ Acc.}
& \makecell{DN-169\\ Top-3\\ Acc.} & \makecell{IN-V3\\ Top-1 \\Acc.}
& \makecell{IN-V3\\ Top-3\\ Acc.} & \makecell{MN\\ Top-1 \\Acc.}
& \makecell{MN\\ Top-3\\ Acc.} & \makecell{SN\\ Top-1 \\ Acc.}\\

 \hline
 $\mathcal{F}_1(x;1.16,1)$ &  90.99  & 98.70  & 90.89  & 98.87 & \textbf{92.14} & 98.99 & \textbf{86.63} & 97.55 & 92.01\\ 
 \hline
 $\mathcal{F}_2(x;1,1)$ &  91.05   & 98.75  & 90.91  & 98.75 & 92.01 & 99.00 & 86.32 & 97.50 & 92.20\\
 \hline 
 $\mathcal{F}_3(x;0.68,1.7)$ &  \textbf{91.12}  & 98.82  & \textbf{90.96}  & 98.79 & 92.12 & 99.05 & 86.00 & 97.36 & \textbf{92.35}\\
 \hline 
 ReLU  &  90.69 & 98.71 & 90.76 & 98.82 & 91.01 & 98.85 & 85.49 & 97.10 & 91.07 \\ 
 \hline
 \makecell{Leaky ReLU \\($\alpha$ = 0.01)} & 90.72 & 98.72 & 90.70 & 98.71 & 91.62 & 98.80 & 85.56 & 97.20 & 91.32\\
 \hline
 \makecell{ELU\\ ($\alpha = 1$)} & 90.31 & 98.41 & 90.55 & 98.70 & 91.09 & 98.69 & 85.59 & 97.24 & 91.01 \\
 \hline
 Swish & 90.80 & 98.85 & 90.70 & 98.75 & 91.50 & 98.81 & 85.55 & 97.15 & 91.70\\
 \hline
 Softplus & 90.55 & 98.75 & 90.31 & 98.42 & 91.59 & 98.72 & 85.54 & 97.10 & 91.01\\
 \hline
  
 \end{tabular}
 \vspace{0.2cm}
\caption{Experimental results on CIFAR10 dataset with different models. ``Acc." stands for Accuracy in \%.} 
\label{tab22}
\end{center}
\end{table}

%............................................
\begin{table}[H]
\begin{center}
\begin{tabular}{ |c|c|c|c|c|c|c|c|c|c|c| }
 \hline
\makecell{Activation\\ Function} &  \makecell{MN\\
Top-1 \\ Acc. }
& \makecell{MN\\ Top-3 \\ Acc.}
& \makecell{IN-V3\\
Top-1\\ Acc. }
& \makecell{IN-V3\\ Top-3\\ Acc.} &  \makecell{DN-121\\
Top-1 \\ Acc.}
& \makecell{DN-121\\ Top-3 \\ Acc.}
& \makecell{DN-169\\
Top-1\\ Acc.}
& \makecell{DN-169\\ Top-3\\ Acc} & \makecell{SN \\ Top-1\\ Acc. } & \makecell{WRN\\ 28-10\\ Top-1 \\Acc.}
\\ 
 \hline
 $\mathcal{F}_1(x;1.16,1)$ &  57.24  & 76.37 & 69.25 & 85.62 &  67.45  &  83.90 & 64.99  & 82.22 & 64.99 & 69.01\\ 
 \hline
 $\mathcal{F}_2(x;1,1)$ &  \textbf{57.60}  &  76.70 &  \textbf{69.31} & 85.65 & \textbf{67.50}  &  83.95 & \textbf{65.78}  & 82.80 & \textbf{65.15} & 69.08\\
 \hline 
 $\mathcal{F}_3(x;0.68,1.7)$ &  57.46  &  76.20 &  69.18 & 85.42 & 67.42  &  83.87 & 65.15  & 82.51 & 65.09 & \textbf{69.20}\\
 \hline
 ReLU  & 56.90 & 76.20 & 69.01  & 85.33 &  66.20 & 83.01 & 64.10 & 81.46 & 62.77 & 66.60\\ 
 \hline
 \makecell{Leaky ReLU\\ ($\alpha$ = 0.01)} & 57.54 & 76.62 & 69.07 & 85.21 & 66.99 & 83.25 & 63.32 & 81.12 & 62.51 & 68.97\\
 \hline
 \makecell{ELU\\ ($\alpha = 1$)}  & 56.99 & 75.95 & 68.55 &  85.14 & 66.62 & 83.49 & 64.32 & 81.54 & 63.60 & 64.56\\
 \hline
 Swish & 56.25 & 75.50 & 68.12 & 84.85 &  66.91 & 83.70 & 64.80 & 82.12 & 64.85 & 68.52\\
 \hline
 SoftPlus & 56.95 & 76.05 & 69.02 & 84.99 &  66.20 & 83.41 & 64.82 & 82.20 & 62.80 & 61.70\\
 \hline
  
 \end{tabular}
 \vspace{0.2cm}
\caption{Experimental results on CIFAR100 dataset with different models. ``Acc." stands for Accuracy in \%.} 
\label{tab23}
\end{center}
\end{table}

%...................................................
\section{Conclusion}
In this paper, we proposed a family of activation functions which we call EIS that involves five hyper-parameters and identified three subfamilies that consistently outperforms well-known activation functions such as ReLU and Swish. Such conclusions were drawn based on experiments carried out on several well-known datasets that provide robust testing grounds. 
%....................................................

%................................................

We also advocate through this article that it is time to move away from simple activation functions and adopt comprehensive search schemes on families of functions to build models. This allows for building more accurate and dependable models. As future work, we will attempt to investigate if this search can be further automated and/or incorporated into the optimization scheme itself. Another scope of future research is to develop a mathematical understanding of reasons leading to improvement in accuracy with perturbation in hyper-parameters in these types of families.

%................................................
\bibliographystyle{unsrt}  
\bibliography{main}  %%% Remove comment to use the external .bib file (using bibtex).
%%% and comment out the ``thebibliography'' section.

%%% Comment out this section when you \bibliography{references} is enabled.
\iffalse

\fi

\end{document}